%% file: main.tex
\documentclass[conference]{IEEEtran}
\IEEEoverridecommandlockouts
\usepackage{cite}
\usepackage{amsmath,amssymb,amsfonts}
\usepackage{algorithmic}
\usepackage{graphicx}
\usepackage{textcomp}
\usepackage{xcolor}
\def\BibTeX{{\rm B\kern-.05em{\sc i\kern-.025em b}\kern-.08em
    T\kern-.1667em\lower.7ex\hbox{E}\kern-.125emX}}

\usepackage{nicefrac}       
\usepackage{hyperref}       
\usepackage{url}            
\usepackage{subcaption}

\usepackage{enumitem}
\usepackage{booktabs}       
\usepackage{microtype}      
\usepackage{multirow}       
\usepackage{orcidlink}      

\title{Neural Velocity for hyperparameter tuning}

\author{
\IEEEauthorblockN{
Gianluca Dalmasso\orcidlink{0009-0005-7354-0838}$^{\dagger}$,
Andrea Bragagnolo\orcidlink{0000-0002-8619-1586}$^{\ddagger}$,
Enzo Tartaglione\orcidlink{0000-0003-4274-8298}$^{\mathsection}$,
Attilio Fiandrotti\orcidlink{0000-0002-9991-6822}$^{\dagger}$,
Marco Grangetto\orcidlink{0000-0002-2709-7864}$^{\dagger}$
}
\IEEEauthorblockA{$^{\dagger}$University of Turin, Turin, Italy}
\IEEEauthorblockA{$^{\ddagger}$LINKS Foundation, Turin, Italy}
\IEEEauthorblockA{$^{\mathsection}$LTCI, Télécom Paris, Institut Polytechnique de Paris, Palaiseau, France}
\IEEEauthorblockA{
name.surname@unito.it,
andrea.bragagnolo@linksfoundation.com,
enzo.tartaglione@telecom-paris.fr
}
}

\begin{document}
\maketitle
\input{abstract}

\input{cap1/_section_1}

\input{cap2/_section_2}

\input{cap3/_section_3}

\input{cap4/_section_4}

\input{cap5/_section_5}

\section*{Acknowledgements}
This paper has been supported by the French National Research Agency (ANR) in the framework of the JCJC project “BANERA” ANR-24-CE23-4369, and by Hi!PARIS Center on Data Analytics and Artificial Intelligence.

\bibliographystyle{IEEEtran}
\bibliography{IEEEabrv, main}

\end{document}

%% file: abstract.tex
\begin{abstract} Hyperparameter tuning, such as learning rate decay and defining a stopping criterion, often relies on monitoring the validation loss. This paper presents NeVe, a dynamic training approach that adjusts the learning rate and defines the stop criterion based on the novel notion of ``neural velocity''.
The neural velocity measures the rate of change of each neuron's transfer function and is an indicator of model convergence: sampling neural velocity can be performed even by forwarding noise in the network, reducing the need for a held-out dataset.
Our findings show the potential of neural velocity as a key metric for optimizing neural network training efficiently.
\end{abstract}

\begin{IEEEkeywords}
Deep learning, Artificial neural networks, Hyper-heuristics, Small data, Neural velocity
\end{IEEEkeywords}

%% file: cap1/_section_1.tex
\input{cap1/1_introduction}

%% file: cap1/1_introduction.tex
\section{Introduction}
\label{sec:1_introduction}

\IEEEPARstart{A}{rtificial} intelligence has made remarkable progress in recent years, driven by the availability of data and powerful deep learning techniques. However, limited training data remains a major challenge in developing robust, generalizable models. This issue is especially critical in small-data scenarios, where acquiring labeled samples is costly or impractical~\cite{9735278, unanue2022regressing}. Training with small datasets often leads to models that memorize specific details rather than learning general patterns, resulting in poor generalization and reduced adaptability to real-world variability. These limitations make deploying such models in dynamic environments particularly risky, highlighting the importance of addressing this challenge.

\begin{figure}[t]
    \centering
    \includegraphics[width=\columnwidth]{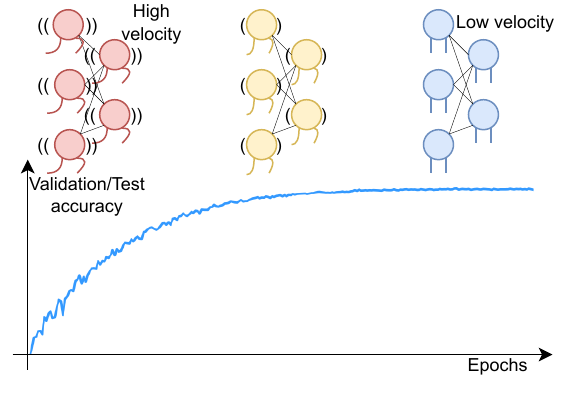}
    \caption{When training a neural network, the neurons in a model change with a progressively decreasing velocity as they reach a plateau in the performance: we can exploit this property to scale the learning rate and to stop the training.}
    \label{fig:teaser}
\end{figure}

Hyperparameter tuning is critical to training successful neural networks, even in scenarios with limited data. Key hyperparameters include the learning rate, the strategy used to adjust it over time, and termination criteria, which determine when to stop training to prevent overfitting and reduce training time. Hyperparameters can be optimized through various methods, ranging from manual tuning based on expert guidance to more sophisticated and computationally intensive techniques like grid search or gradient-based approaches~\cite{Chandra2022GD, cutkosky2023mechaniclearningratetuner}, which automate the process. However, most tuning methods require holding out a \textit{validation} set from the training data, which presents notable challenges. With limited annotated data, reserving samples for validation reduces the data available for training, potentially impairing the model's learning ability. Furthermore, validation sets may not accurately reflect real-world scenarios, leading to misleading evaluations~\cite{lutz1998automatic}. To address these issues, there is growing interest in validation-free learning approaches, which aim to optimize hyperparameters without relying on separate validation samples, offering a promising alternative in data-constrained environments.

Another critical component of training successful neural networks is the stopping criteria. Early stopping is a common technique that halts training when a model’s performance on a validation set stops improving, often using a patience parameter to define the number of epochs without improvement~\cite{Prechelt2012early}. Advanced methods, such as that in~\cite{yao2007early}, refine this process by adapting stopping criteria based on training dynamics, improving efficiency and generalization. However, these methods rely on a validation set, which can lead to biased decisions if it does not represent the broader data distribution accurately. More recent approaches leverage noisy labels~\cite{yuan2024early}, avoiding the use of validation data by tracking changes in the model’s predictions during training. These methods determine the optimal stopping point by detecting when the model starts overfitting mislabeled data, thus improving robustness in noisy settings.

This paper introduces NeVe, a novel method to optimize learning rate decay and stopping criteria in neural network training without requiring a held-out validation set. NeVe leverages the concept of neural velocity, defined as the rate at which neurons change their input-output behavior during training. The method dynamically adjusts the learning rate when a neuron's change stabilizes, signaling it has reached a ``meta-stable'' state (Fig.~\ref{fig:teaser}). Training is terminated when neural velocity approaches zero, indicating convergence to a stable solution. By utilizing an auxiliary dataset of random samples, NeVe not only minimizes reliance on validation data but also addresses the inefficiencies that arise in data-scarce scenarios, enabling a more efficient and adaptive training procedure while preserving valuable training data for model optimization.

The main contributions and findings presented in this paper are summarized in the following key points:
\begin{itemize}[noitemsep, nolistsep]
    \item \textbf{First Validation-less Approach Using Neural Velocity}: NeVe is, to the best of our knowledge, the first method to use the concept of neural velocity to both scale the learning rate decay and provide a stopping criterion without a validation set (Sec.~\ref{sec:3_neve_estimation}).
    \item \textbf{Plug-and-Play Strategy}: Unlike other methods, NeVe is a versatile, plug-and-play approach that can automatically adapt to different architectures, datasets, and optimizers (Sec.~\ref{sec:3_neve_estimation}, Sec.~\ref{sec:4_ablation}).
\end{itemize}

This paper is organized as follows. In Section~\ref{sec:2_related_works}, we review related works, providing a discussion of their strengths and weaknesses. Section~\ref{sec:3_method} focuses on our proposed technique, the algorithm, and an overview of its key hyperparameters. Section~\ref{sec:4_exps} presents experimental results, detailing the characteristics of our method, the experimental setup, and a thorough analysis of the results, including an ablation study. Finally, in Section~\ref{sec:5_conclusion}, we summarize our findings, revisit the proposed technique, and discuss the broader implications of this work.

%% file: cap2/_section_2.tex
\input{cap2/2_related_works}

%% file: cap2/2_related_works.tex
\section{Related Works}
\label{sec:2_related_works}

In this section, we provide a primer on hyperparameter tuning strategies, learning rate decay policy, and the stopping criteria that are especially relevant to this work.

Hyperparameter tuning is a widely studied topic in machine learning, focusing on finding the optimal configuration to maximize model performance. While grid search~\cite{hutter2011sequential, snoek2012practical} provides an exhaustive exploration of the hyperparameter space, it becomes computationally expensive in high-dimensional settings due to the curse of dimensionality. Alternative methods, such as Bayesian optimization~\cite{mockus2005bayesian}, use probabilistic models to map hyperparameters to objectives, achieving better results than grid or random search. Evolutionary algorithms~\cite{kim2019evolution, wieser2020eo} offer another approach, iteratively improving a population of hyperparameter configurations by selecting and refining the fittest candidates. Gradient-based techniques~\cite{maclaurin2015gradient, Chandra2022GD, li2023scotti}, particularly suited for neural networks, enable direct optimization of hyperparameters, such as the learning rate and early stopping thresholds, alongside model parameters. These methods reduce sensitivity to initial choices and automate hyperparameter adjustment during training.

The learning rate is often regarded as the most influential hyperparameter~\cite{GoodBengCour16}, and unlike others, it is rarely kept constant during training. Often it is dynamically adapted during training to improve convergence and generalization~\cite{li2019towards}. Fixed learning rates are rarely effective, as high initial rates help explore the loss landscape, while lower rates stabilize convergence. Strategies such as step-wise decay, exponential decay, or validation-based adjustments~\cite{ackley1985learning, Krizhevsky2012alexnet, simonyan2015vgg} are commonly used to control the learning rate over time. Sophisticated learning algorithms with adaptive learning rates have been proposed~\cite{kingma2015adam}, but even with these methodologies, a learning rate decay policy is still applied~\cite{dosovitskiy2021an}.

Early stopping is a widespread technique used to limit the overfitting of neural networks and lessen the training cost, thanks to early termination of the training process~\cite{finnoff1993improving, lodwich2009evaluation}. However, validation signals are often noisy, with multiple local minima complicating decision-making (i.e., performance may still improve after it has begun to decrease)~\cite{baldi1991temporal, dodier1995geometry}. Robust criteria, such as generalization loss (e.g., stopping when validation error increases by a threshold) or uninterrupted progress (e.g., halting after consecutive validation increases)~\cite{lutz1998automatic}, aim to address these challenges. Despite their effectiveness, these methods depend on held-out validation sets, which reduce the available training data and exacerbate overfitting risks in high-variance datasets. We work around such a problem by introducing neural velocity to measure a neuron's trend of ``change'' during training. Neural velocity represents a way to estimate the learning dynamics, which avoids sacrificing some data for validation to control the learning rate policy and early stopping strategy.

%% file: cap3/_section_3.tex
\input{cap3/3_method}

%% file: cap3/3_method.tex
\section{Proposed Method}
\label{sec:3_method}

In this section, we introduce the core concept of neural velocity as an estimator of the rate of change of a neuron input-output function (Sec.~\ref{sec:3_neve_estimation}).
We make no assumptions about the architecture topology or the nature of the learning problem. We then explain (Sec.~\ref{sec:3_neve_hyp_estimation}) its most critical hyperparameter.

\subsection{Neural Velocity Estimation}
\label{sec:3_neve_estimation}
This section's main contribution is the notion of neural velocity, which builds on the concept of a neuron's rate of change.
Let us define some generic auxiliary dataset $\mathcal{D}_{\text{aux}}$ that does not necessarily coincide with the typical validation set $\mathcal{D}_{\text{val}}$.
We evaluate the rate of change for the $i$-th neuron by using a sampling strategy on $\mathcal{D}_{\text{aux}}$.
For a sample $\xi \in \mathcal{D}_{\text{aux}}$, the $i$-th neuron at time $t$ will produce the output $\boldsymbol{y}_{i,\xi}^t$ which is then normalized, obtaining $\hat{\boldsymbol{y}}_{i,\xi}^t$. Similarly, we can estimate the neuron change rate as:

\begin{equation}
    \hat{\rho}^t_i = \left<\hat{\boldsymbol{y}}_{i,\mathcal{D}_{\text{aux}}}^t, \hat{\boldsymbol{y}}_{i,\mathcal{D}_{\text{aux}}}^{t-1}\right>.
    \label{eq:changerate}
\end{equation}

\begin{figure}[t]
    \begin{subfigure}{0.40\columnwidth}
        \hfill
        \includegraphics[width=\columnwidth, trim=0 0 0 0, clip]{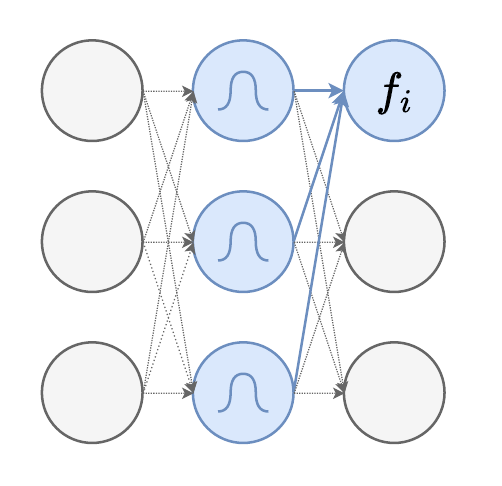} 
        \caption{~}
        \label{fig:randinput}
    \end{subfigure}
    ~
    \begin{subfigure}{0.57\columnwidth}
        \hfill
        \includegraphics[width=\columnwidth, trim=0 -7 0 0, clip]{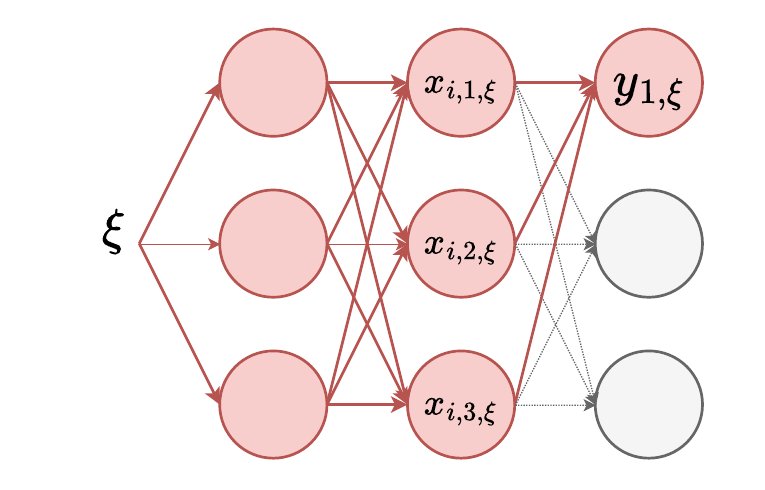} 
        \caption{~}
        \label{fig:samplinginput}
    \end{subfigure}
    \caption{Sampling of the function given random Gaussian variables (a), and direct sampling from input patterns $\xi$ (b).}
    \label{fig:chap3_training_scheme}
\end{figure}

The computation of $\hat{\rho}^t_i$ over the full $\mathcal{D}_{\text{aux}}$ follows~\cite{bragagnolo2022update}. Such formulation provides a practical advantage over an analytical model: 
the input domain of the $i$-th neuron dynamically varies across the epochs: despite having the same $\mathcal{\xi}$, it is possible that $\boldsymbol{x}_i^{t}\neq \boldsymbol{x}_i^{t-1}$ and that $\boldsymbol{w}_i^{t}\neq \boldsymbol{w}_i^{t-1}$, meaning that the evaluated function is not limited to the input of the $i$-th neuron, but it is the function from the input of the neural network to the $i$-th neuron (Fig.~\ref{fig:samplinginput}). This means that all the internal correlations and permutations (which could occur at training time) are taken into account through a simple forward propagation step, which is not done assuming random Gaussian inputs of the single target neuron (Fig.~\ref{fig:randinput}). Given that $\hat{\rho}^t_i$ estimates the change rate of the $i$-th neuron between $t$ and $t-1$, to determine its smoothened evolution, we define the \emph{neural velocity} of the $i$-th neuron as:
\begin{equation}
    \centering
    \label{eq:cap3_velocity}
    v^t_i = \left|(1 - \hat{\rho}^t_i) - \mu_{\text{vel}}v^{t-1}_i\right|,
\end{equation}

where $\mu_{\text{vel}}$ is a smoothening hyperparameter. The neural velocity tells us how fast the neuron changes its output (with respect to the previous epoch) when the model receives the same input. Intuitively, when this value is high, the $i$-th neuron (and the underlying graph between itself and the input) is still changing the learned function $f_i$ and has not reached convergence yet. In contrast, when the neural velocity has settled close to zero, it has reached convergence despite it is potentially still changing parameters.

We take advantage of the notion of neural velocity to adjust the learning rate as follows.
Let us define the model velocity as the average of the velocities of all the neurons in the model.
It is well known that at high learning rates, performance plateaus may arise due to induced stochastic noise~\cite{liu2021noise}.
Similarly, we can expect the model velocity to reach a plateau when the model is in these meta-stable states.
Towards this end, a velocity plateau allows detecting and coping with performance plateaux early in the learning process by lowering the learning rate, as we discuss in the following.

\begin{figure}[t]
    \centering
    \includegraphics[width=0.75\columnwidth, trim=0 0 0 0, clip]{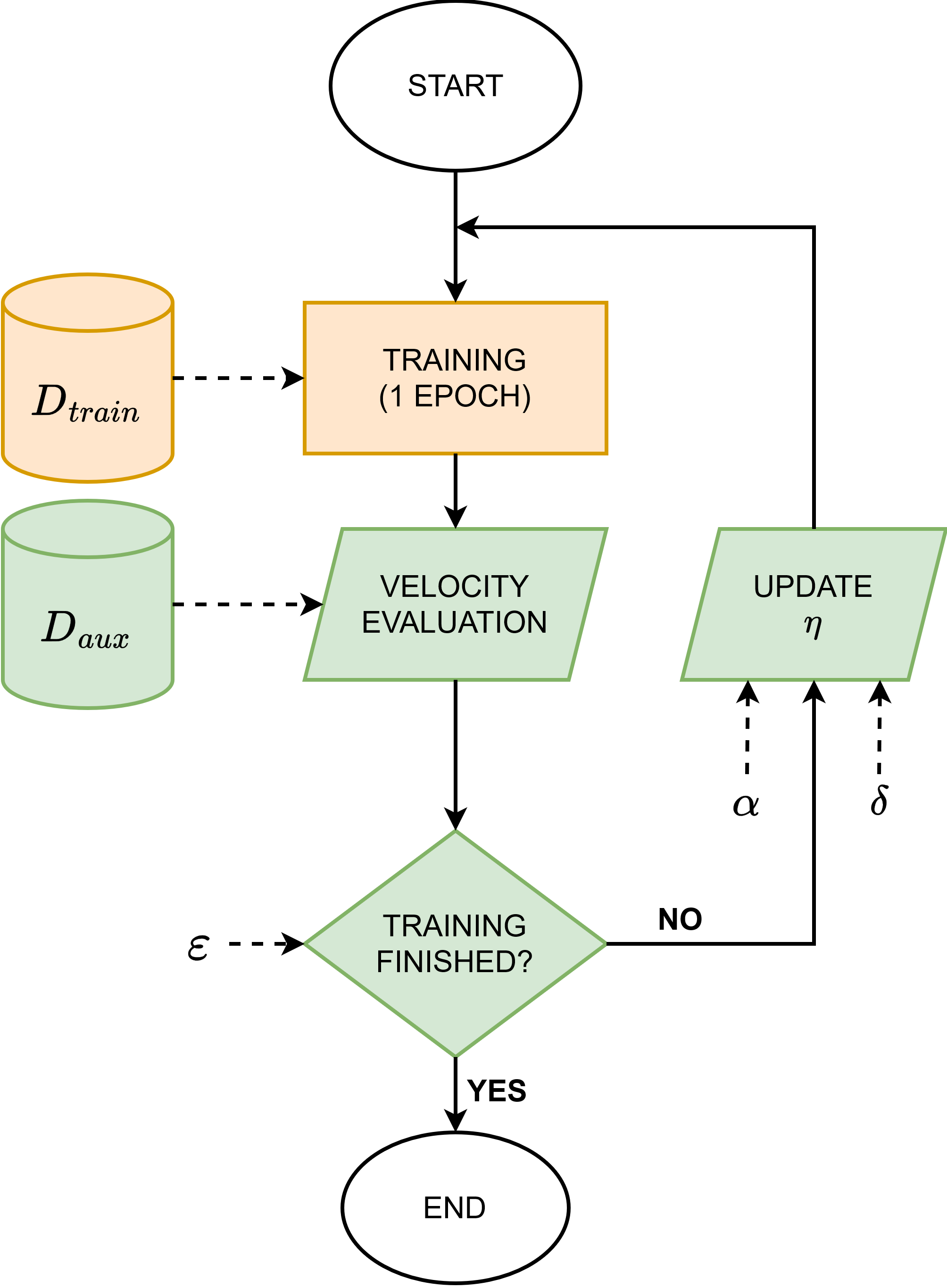}
    \caption{Overview of NeVe. After each training epoch, we perform a velocity evaluation: if the velocity is under a certain threshold $\varepsilon$, we stop the training; otherwise, we may update the learning rate $\eta$.}
    \label{fig:training_scheme}
\end{figure}

To address this, we introduce a training procedure built around the concept of model velocity, as illustrated in Fig.~\ref{fig:training_scheme}. Let the auxiliary set $\mathcal{D}_{\text{aux}}$ be a set of samples distinct from the training set that will be used to calculate the velocities of the neurons. Given a randomly initialized or pre-trained model, the model velocity must be computed before the first training epoch has elapsed. Towards this end, we evaluate $\hat{\boldsymbol{y}}_{i, \mathcal{D}_{\text{aux}}}^{t=0}$, i.e., the normalized output for each neuron in the model. After the first training epoch, we calculate the first change rates $\hat{\rho}_i^{t=1}$ according to \eqref{eq:changerate}. Assuming $v_{i, \mathcal{D}_{\text{aux}}}^{t=0}=0$, we can evaluate $v_{i, \mathcal{D}_{\text{aux}}}^{t=1}$.

Once the velocity of all neurons is computed, their average is taken as the model velocity. When this velocity stabilizes to some value $k$ for $\delta$ epochs of patience, the model is considered to have reached a meta-stable state, and the learning rate is rescaled by a factor $\alpha$. As training progresses, if the model velocity drops below a threshold $\varepsilon$, the training is considered to be concluded, indicating that the model has reached convergence. As suggested in~\cite{bragagnolo2022update}, we set $\mu_{\text{vel}}=0.5$ to prevent excessive velocity fluctuations for ReLU neurons.


\subsection{Choice of  \texorpdfstring{$\varepsilon$}{ε}}
\label{sec:3_neve_hyp_estimation}
Let us assume for the sake of simplicity that $\mu=0$. This means that the velocity is evaluated as $v^t=1-\hat{\rho}^t$. When we threshold by $\varepsilon$, we are essentially checking that between two epochs the output of the neurons, in the average case, will be $\Delta y_i^t= |y_i^t - y_i^{t-1}| \leq \varepsilon + M_i^t$, where $M_i^t$ is proportional to the variation in the average activation intensity for the $i$-th neuron. Let us assume $M_i=0$ and consider the case in which we are at the output (softmax-activated) layer, where the velocity is exactly $\varepsilon$. We write the average variation of the output for the whole model as: 
\begin{align}
    \Delta y_i^t &= \frac{(e^{z_i})^{1-\varepsilon}}{\sum_j(e^{z_j})^{1-\varepsilon}} - \frac{e^{z_i}}{\sum_j e^{z_j}} \nonumber\\&= \frac{e^{z_i}}{\sum_j e^{z_j}} \left[\left(\frac{e^{z_i}}{\sum_je^{z_j}}\right)^{-\varepsilon} - 1\right] = p_i(p_i^{-\varepsilon} - 1),
\end{align}
where $z_i$ is the output of the $i$-th neuron before passing through the softmax, and $p_i=\frac{e^{z_i}}{\sum_j e^{z_j}}$. Here, we want to find what is the $p_i$ such that we induce the maximum $\Delta y_i$. 

Considering $p_i\in (0, 1)$, we find that the absolute maximum is for $p_i=(1-\varepsilon)^{\frac{1}{\varepsilon}}$. If we substitute back, we see that the maximum expected variation at the output of the softmax is
\begin{equation}
    \max\{\Delta y_i\} = (1-\varepsilon)^{\frac{1}{\varepsilon}}[(1-\varepsilon)^{-1} -1].
\end{equation}
When employing ${\varepsilon=10^{-3}}$, for example, we have $\max\{\Delta y_i\}\approx 3.7\times 10^{-4}$: in case the number of classes is 1k (hence, the $p_i$ for the correct class should be larger than $10^{-3}$), this choice of $\varepsilon$ is safe for indicating stability along the training process. We will ablate over different choices of $\varepsilon$ in Sec.~\ref{sec:4_ablation}.

%% file: cap4/_section_4.tex
\input{cap4/4_results}

%% file: cap4/4_results.tex
\section{Experimental Results}
\label{sec:4_exps}
In this section, we experiment with NeVe on different computer vision datasets. For each, we compare NeVe with the state-of-the-art training procedure.
Additionally, we examined a variant technique that uses a loss value estimated on an actual validation set instead of the velocity to simulate a naive version of our procedure. All the experiments were conducted using NVIDIA A40 GPUs and PyTorch~1.13.1. Open source code can be found in Github.~\footnote{\url{https://github.com/EIDOSLAB/NeuralVelocity}}

\subsection{Defining the Auxiliary Dataset}
\begin{figure}[t]
    \centering
    \begin{subfigure}{\columnwidth}
        \centering
        \includegraphics[width=0.9\columnwidth]{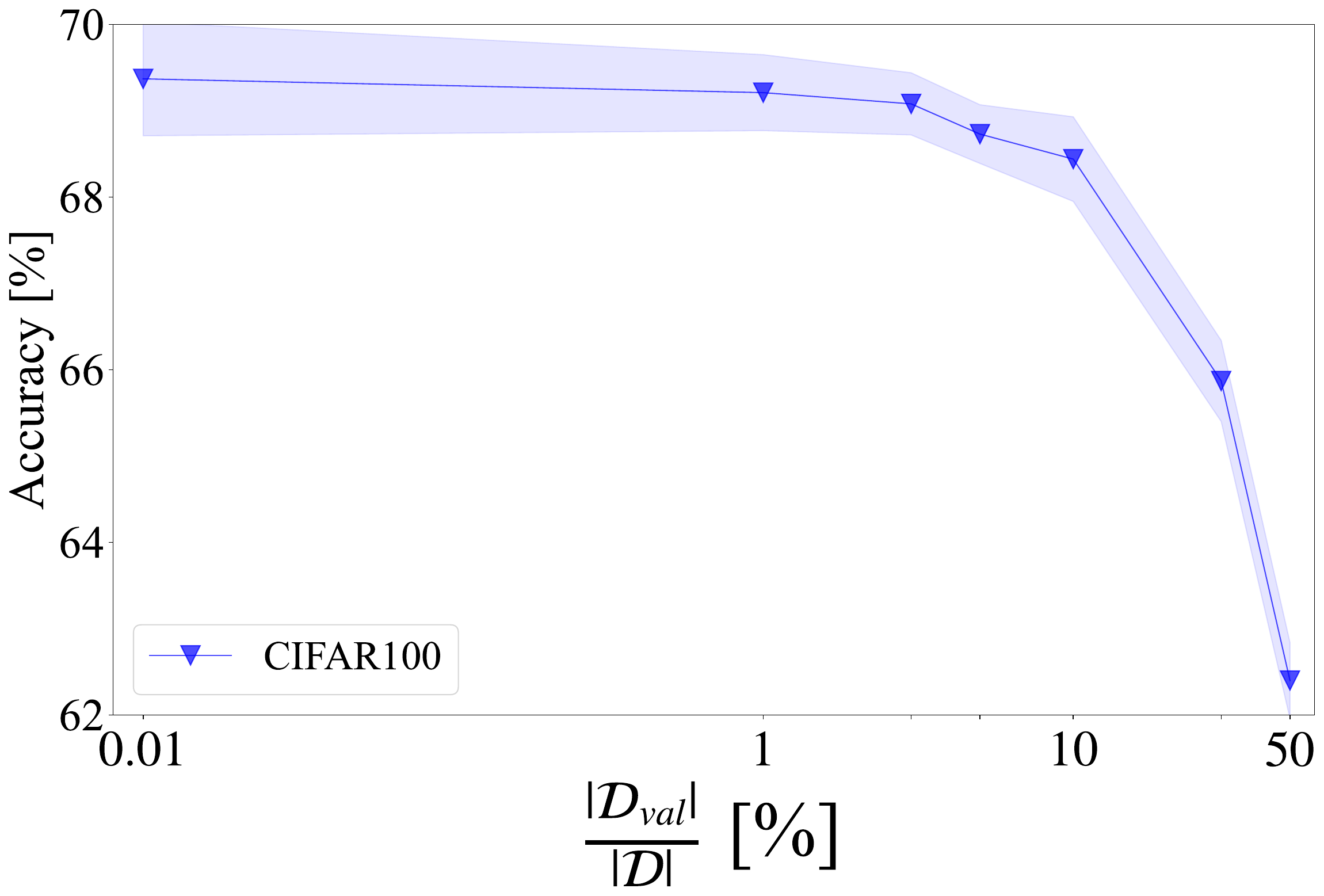}
        \caption{~}
        \label{fig:cap4_ablation_aux_dataset_2}
    \end{subfigure}
    ~
    \begin{subfigure}{\columnwidth}
        \centering
        \includegraphics[width=0.9\columnwidth]{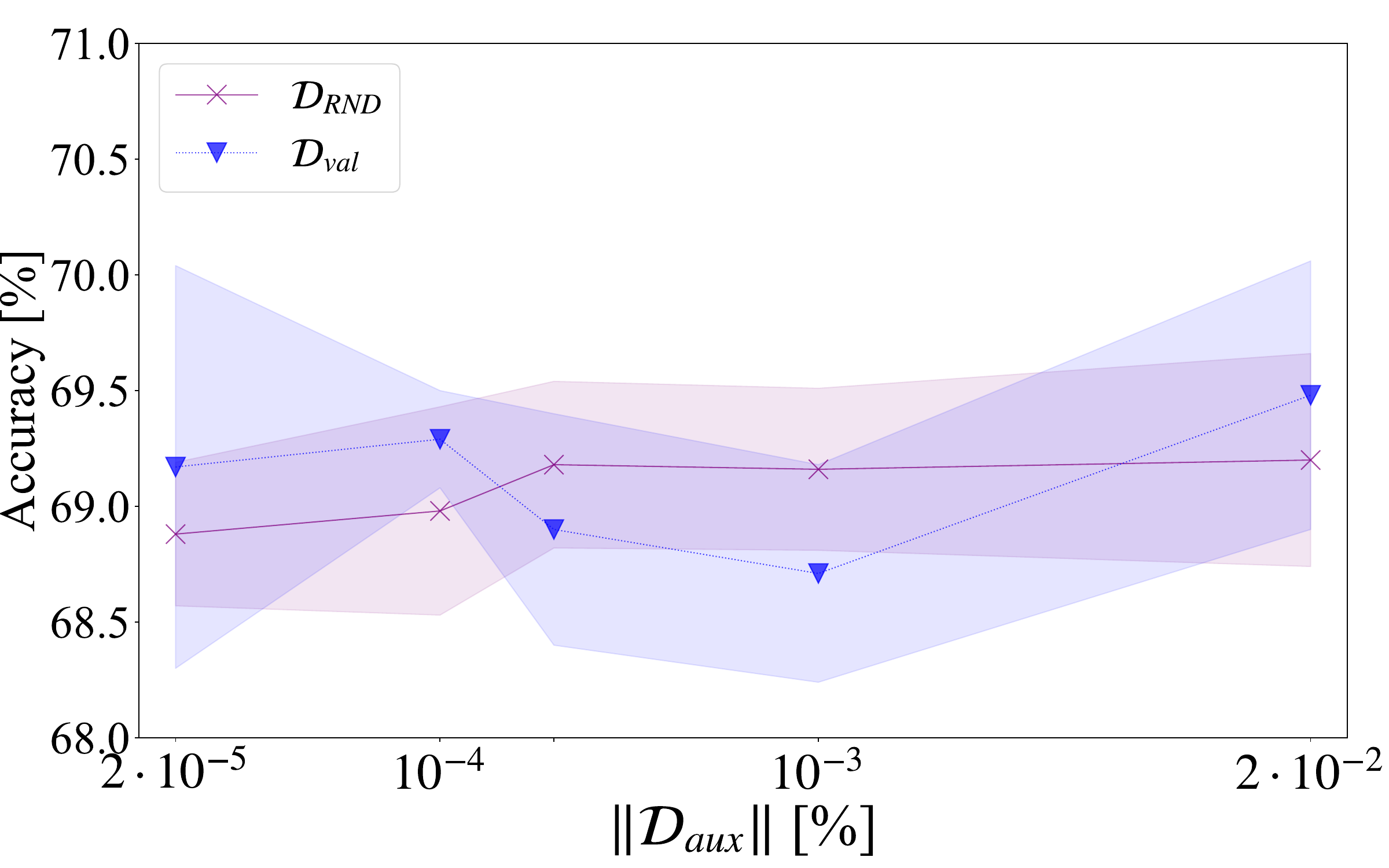}
        \caption{~}
        \label{fig:cap4_ablation_aux_dataset_1}
    \end{subfigure}
    \caption{Effect of changing $\mathcal{D}_\text{val}$ and $\mathcal{D}_\text{aux}$ relative size for a ResNet32 classifier on CIFAR100: test accuracy as a function of the number of images moved from the training to the validation set (a), test accuracy as a function of the auxiliary set (b).}
    \label{fig:cap4_ablation_aux_dataset}
\end{figure}

First, we analyze the impact of holding out images from the training set to create a validation set. We train a ResNet-32~\cite{he2016deep} following the reference setup in~\cite{BMVC2016_87} on CIFAR100~\cite{cifar100}. Fig.~\ref{fig:cap4_ablation_aux_dataset_2} shows that withholding just 10\% of the training samples for validation already degrades performance, highlighting the need for a validation-free approach like NeVe. Interestingly, as observed in~Fig.~\ref{fig:cap4_ablation_aux_dataset_1}, computing model velocity on random noise yields results comparable to using a validation set, even when considering a small size for $\mathcal{D}_{\text{RND}}$. This allows NeVe to rely on a small auxiliary dataset, thereby reducing complexity. Consequently, in the remainder of our experiments, we use an auxiliary set of 100 images generated from random Gaussian noise due to its simplicity.

\subsection{Experimental Setup}
\label{sec:experimentres}
In this section, we validate NeVe on three image classification datasets. We compare our approach with each setup’s state-of-the-art baseline (``Baseline''). In addition, we consider a validation loss-based approach (``V.Loss''), using 10\% and 30\% of the original training set as validation data, where the validation loss is used in place of the neural velocity term.

Following our previous experiments, we upscale the complexity of the classification task to ImageNet-100, a 100-classes subset of the ImageNet-1k dataset~\cite{ILSVRC15}. For this experiment, we also scale the classifier complexity and train both a ResNet-50 convolutional network and the SwinT-v2 Transformer~\cite{liu2022swin}. We used state-of-the-art procedures and hyperparameter search methods to build a baseline performance.

The ResNet-32 was trained using the setup in \cite{BMVC2016_87}. For the SwinT-v2 Transformer, we used AdamW with the default hyperparameters defined by the authors,~\footnote{\url{https://github.com/microsoft/Swin-Transformer}} while for ResNet-50 we used SGD with $\eta=0.1$, momentum $0.9$ and weight decay $10^{-4}$ using a batch size of $128$. In these (baseline) experiments, we do not use validation loss to assess the learning process or perform any train-validation split. Instead, these experiments employed the reported hand-crafted hyperparameters that had previously proven effective in those configurations.

\textbf{\textit{Data Augmentation}} To mitigate overfitting, we applied data augmentation in all our experiments, tailoring transformations to each dataset. For CIFAR-10 and CIFAR-100, we used a RandomCrop (32×32) with a padding of 4, followed by a RandomHorizontalFlip and normalization. For ImageNet, images were resized to 232×232, then randomly cropped to 224×224, followed by a RandomHorizontalFlip and normalization.

\subsection{Discussion}
The results of our experiments, shown in Tab.~\ref{tab:cap4_experiments}, report mean and variance over 10 runs and demonstrate that NeVe achieves state-of-the-art performance by self-tuning the learning rate without requiring a validation set.

\begin{figure}[t]
    \centering 
    \includegraphics[width=0.9\columnwidth]{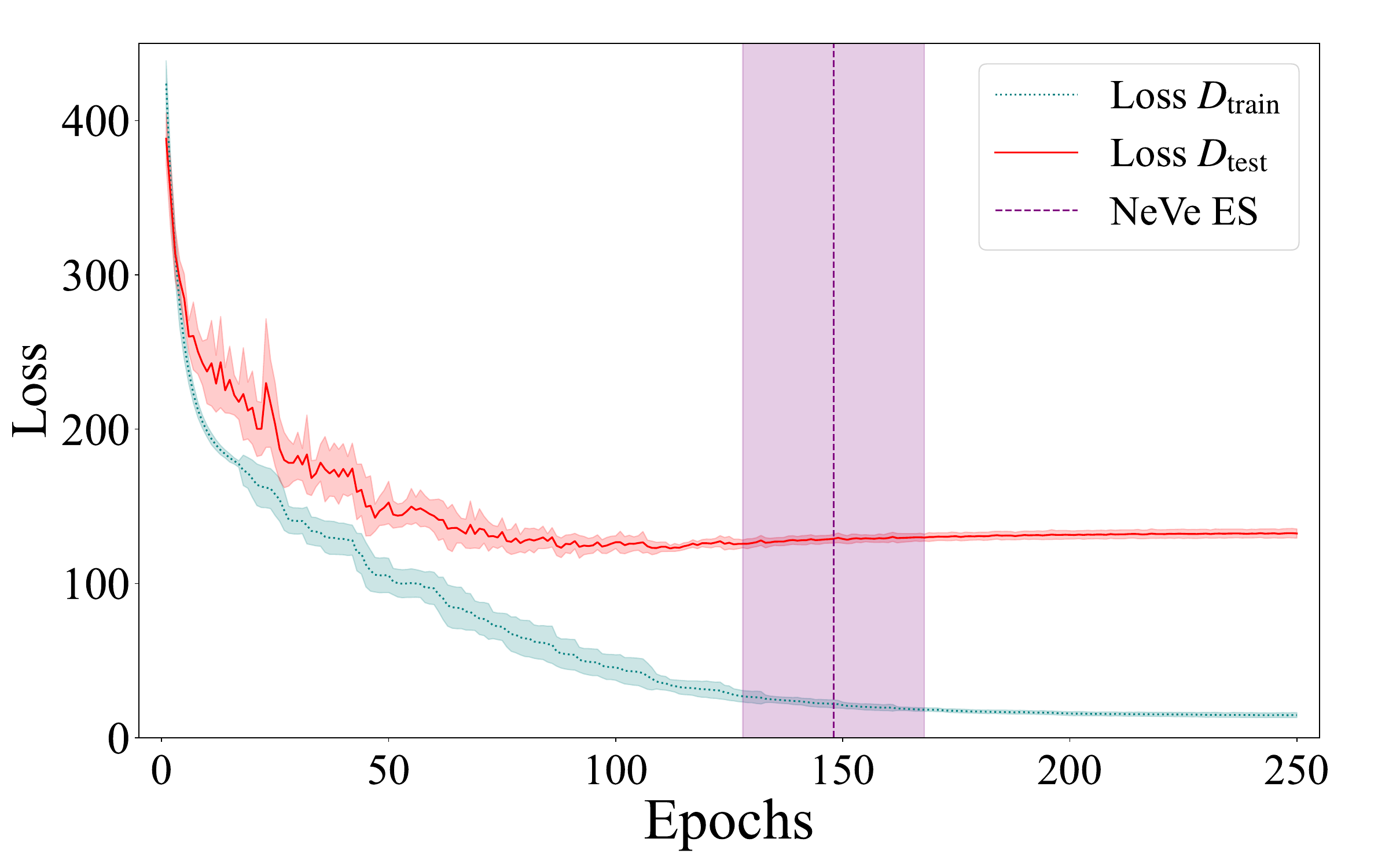}
    \caption{NeVe stop proposal on CIFAR100, the vertical purple line represents the mean and std of the proposed stop epochs evaluated over 10 seeds/runs.}
    \label{fig:rebuttal2_neve_es_with_losses_cifar100}
\end{figure}

A key factor behind NeVe’s effectiveness is its stopping mechanism, which detects the beginning of overfitting and stops training at an optimal point (as shown in Fig.~\ref{fig:rebuttal2_neve_es_with_losses_cifar100}). While this approach resembles traditional early stopping—where training stops when the test loss starts increasing while the training loss still decreases—NeVe differs by not relying on validation loss monitoring. Instead, it operates solely on an auxiliary set, $D_{\text{aux}}$, maximizing data utilization while mitigating overfitting. This leads to improved generalization across various tasks.

\textbf{\textit{Computational Costs}} NeVe requires evaluating the state of all the neurons of the network (e.g., through PyTorch forward hooks) over $|D_{\text{aux}}|$ samples. If the neuron's output size is $N_i$, we have a space complexity of $\mathcal{O}(|D_{\text{aux}}| \cdot N_i)$ for each neuron, which, given a small $|D_{\text{aux}}|$ (in Fig.~\ref{fig:cap4_ablation_aux_dataset_1} we show that we need very few samples), is negligible. For what concerns memory complexity, NeVe requires storing the state of all the neurons once before the training starts (so that we can evaluate the velocity from the first epoch). We have to do so for $|D_{\text{aux}}|$ samples, this give us a complexity of $\mathcal{O}(|D_{\text{aux}}|)$ for each neuron. If we train the model for L epochs (for simplicity, we do not consider early-stop), we must evaluate the velocity L times. To evaluate the velocity, we first need to store the updated neuron state, we then need to evaluate $\hat{\rho}^{t}$, and finally we can evaluate $v^{t}$. The complexity of $\hat{\rho}^{t}$ is $\mathcal{O}(|D_{\text{aux}}|^2)$ since we evaluate the similarity between the output of a neuron in two different time-step. The complexity to evaluate $v^{t}$ is $\mathcal{O}(1)$ since it is a simple sum of values. Finally, the time complexity (TC) for training over $L$ epochs and $M$ neurons is:
\begin{align*}
    TC &= \mathcal{O}(M \cdot |D_{\text{aux}}|) + \mathcal{O}(L \cdot |D_{\text{aux}}|) + \mathcal{O}(|D_{\text{aux}}|^2) + \mathcal{O}(1)
    \nonumber\\&= \mathcal{O}( M \cdot L \cdot |D_{\text{aux}}|^2)
\end{align*}
which again, for a small $|D_{\text{aux}}|$, is negligible.

\begin{figure}[t]
    \centering
    \begin{subfigure}{\columnwidth}
        \centering
        \includegraphics[width=0.9\columnwidth]{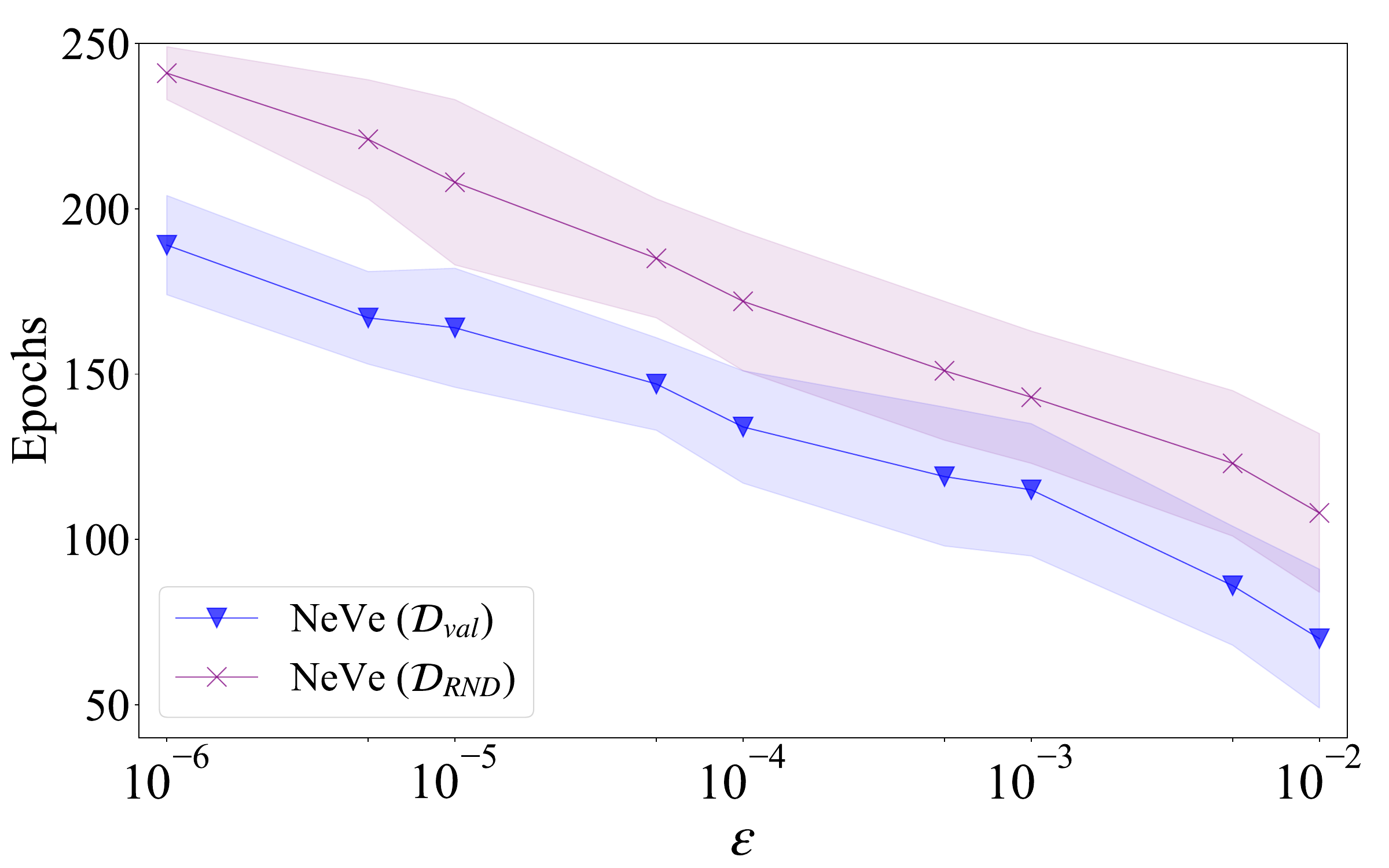}
        \caption{~}
        \label{fig:cap4_ablation_epsilon_epochs}
    \end{subfigure}
    ~
    \begin{subfigure}{\columnwidth}
        \centering
        \includegraphics[width=0.9\columnwidth]{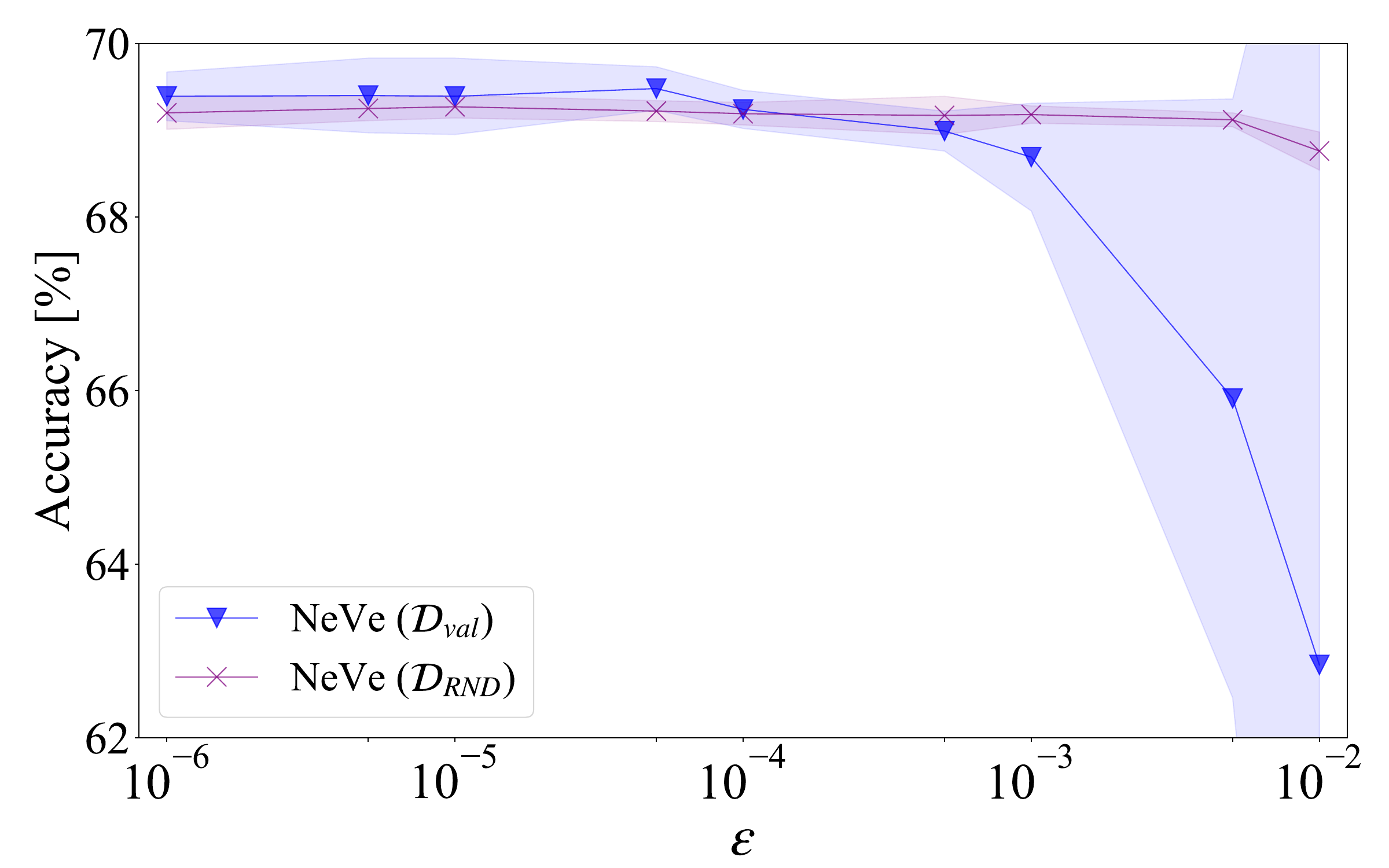}
        \caption{~}
        \label{fig:cap4_ablation_epsilon_accuracy}
    \end{subfigure}
    \caption{NeVe's sensitivity to velocity threshold $\varepsilon$ on CIFAR100: number of training epochs (a), and test set accuracy (b).}
    \label{fig:cap3_ablation_epsilon}
\end{figure}

\textbf{\textit{NeVe Hyperparameters Tuning}}
The hyperparameter tuning of our proposed technique involves three key parameters: $\varepsilon$, $\alpha$, and $\delta$. The parameter $\varepsilon$ is critical for the early stopping mechanism; specifically, a smaller value of $\varepsilon$ results in the early stopping operation being triggered later in the training process, which occurs when the model's output exhibits minimal changes. This relationship is illustrated in Fig.~\ref{fig:cap3_ablation_epsilon}, where a larger value of $\varepsilon$ prompts early stopping at an earlier stage, potentially preventing the model from reaching convergence. The parameters $\alpha$ and $\delta$ further influence the overall learning dynamics of the model. Specifically, the parameter $\alpha$ acts as a rescaling factor for the learning rate; lower values of $\alpha$ lead to a more pronounced reduction in the learning rate with each update, which results in fewer, more conservative changes to the model's weights during backpropagation.
In contrast, higher values of $\alpha$ result in smaller reductions in the learning rate, allowing more substantial weight updates and faster shifts in the optimization process. Consequently, excessively low and high values of $\alpha$ may lead to suboptimal convergence. Similarly, $\delta$ controls how frequently the learning rate is updated, a lower value causes frequent changes, increasing the likelihood of converging to a poorer local minimum. On the other hand, a higher $\delta$ leads to less frequent updates, potentially causing an insufficient decrease in the learning rate, which can ultimately hinder the model’s convergence to an optimal solution.

\begin{figure}[t]
    \centering
    \begin{subfigure}{\columnwidth}  
        \centering 
        \includegraphics[width=0.9\columnwidth]{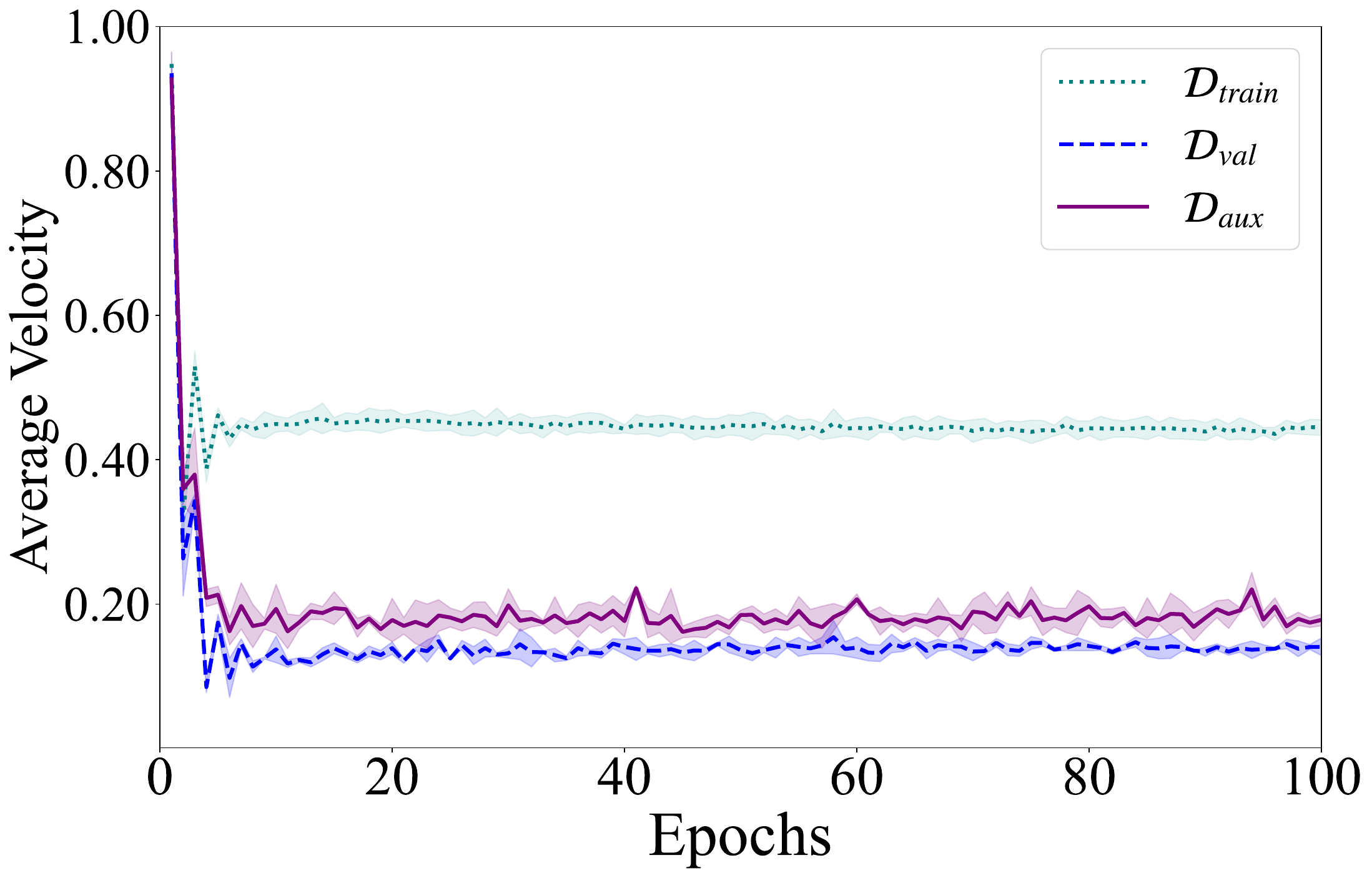}
        \caption{~}
        \label{fig:rebuttal_velocity_evaluated_daux_avg_velocity}
    \end{subfigure}
    ~
    \begin{subfigure}{\columnwidth}  
        \centering 
        \includegraphics[width=0.9\columnwidth]{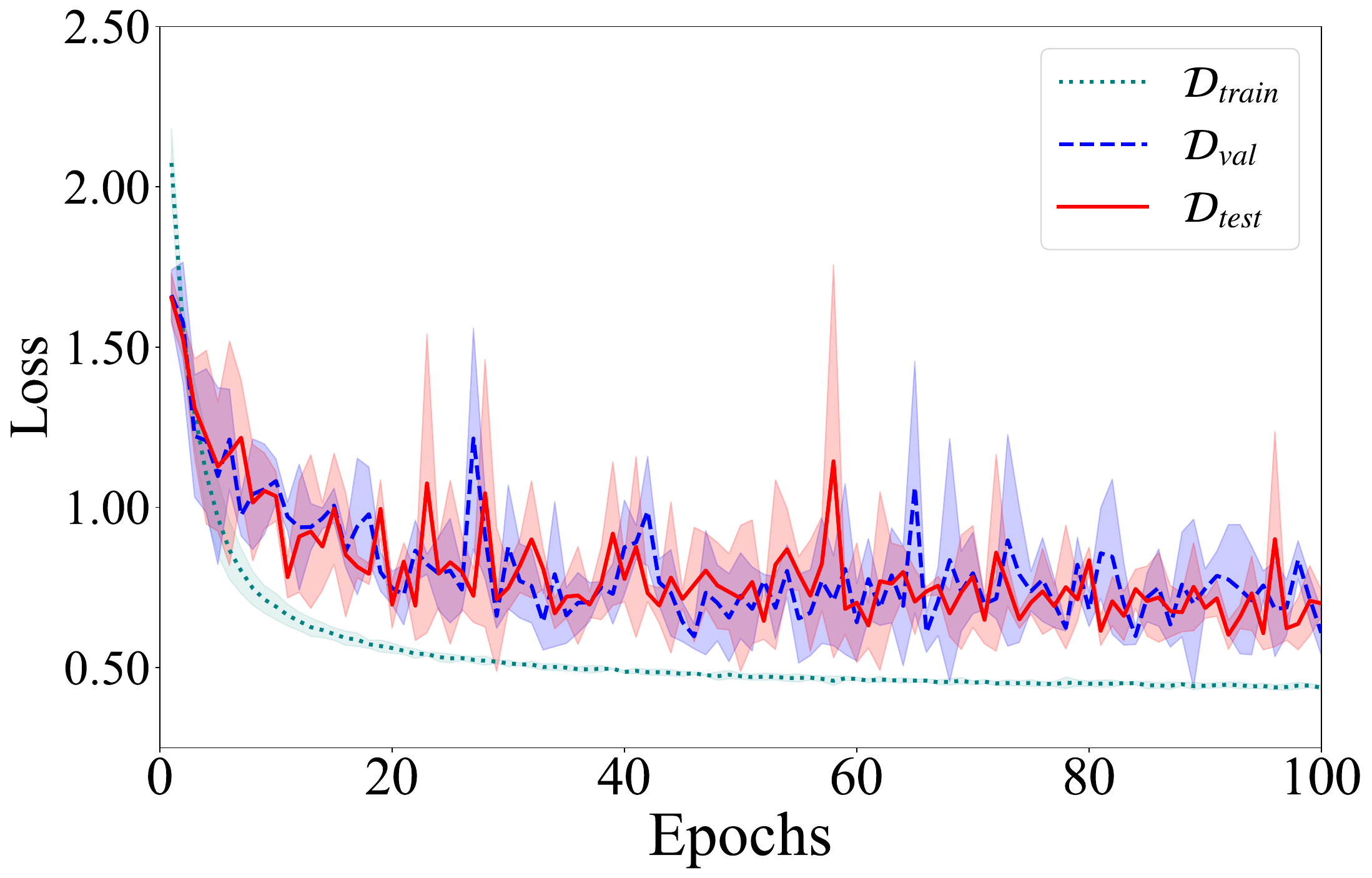}
        \caption{~}\label{fig:rebuttal_velocity_evaluated_daux_loss}
    \end{subfigure}
    \caption{Experiments conducted on CIFAR10 with no $\eta$-scheduler used: average velocity evaluated on different datasets (train, validation, and auxiliary) (a), and train, validation, and test losses (b).}
    \label{fig:rebuttal_velocity_evaluated_on_different_daux}
\end{figure}

\textbf{\textit{Correlation Velocity - Loss}}
Fig.~\ref{fig:rebuttal_velocity_evaluated_daux_avg_velocity} compares the velocity evaluated on three datasets: the training, validation, and auxiliary sets. Here, we see that the behavior of the velocity computed on the auxiliary set closely matches the velocity computed on the validation set. Fig.~\ref{fig:rebuttal_velocity_evaluated_daux_loss} shows the same experiment's train, validation, and test losses. We can see that after 20 epochs when we find the beginning of a plateau in the velocity, the validation loss stabilizes; this shows a trend correlation between the velocity and the validation loss.

\begin{table}[t]
    \centering
    \caption{Comparison with related works. Experiments were conducted with a ResNet-32 model over 10 runs.}
    \label{tab:comparison_related_works}
    \resizebox{0.45\textwidth}{!}{
    \begin{tabular}{c c c c}
        \toprule
        \bf Dataset & \bf Method & \bf Patience &\bf Test Accuracy [\%] \\
        \midrule
        \multirow{6}{*}{CIFAR10} & Ultimate Optimizer & - & $86.77 \pm 0.78$ \\
        \cmidrule{2-4}
        & \multirow{2}{*}{Label Wave} & 5 & $ 61.65 \pm 8.06 $ \\
        & & 10 & $ 63.65 \pm 8.30 $ \\
        \cmidrule{2-4}
        & \multirow{2}{*}{\bf NeVe} & \bf 5 & $\mathbf{92.49 \pm 0.40}$ \\
        & & \bf 10 & $\mathbf{93.10 \pm 0.30}$ \\
        \midrule
        \multirow{6}{*}{CIFAR100} & Ultimate Optimizer & - & $57.89 \pm 1.52$ \\
        \cmidrule{2-4}
        & \multirow{2}{*}{Label Wave} & 5 & $ 33.30 \pm 2.77 $ \\
        & & 10 & $ 33.81 \pm 2.82 $ \\
        \cmidrule{2-4}
        & \multirow{2}{*}{\bf NeVe} & \bf 5 & $\mathbf{68.90 \pm 0.61}$ \\
        & & \bf 10 & $\mathbf{69.18 \pm 0.10}$ \\
        \bottomrule
    \end{tabular}
    }
\end{table}

\textbf{\textit{Comparison with Related Works}} In Sec.~\ref{sec:2_related_works}, we introduced two recent algorithms for hyperparameter tuning: Gradient Descent: The Ultimate Optimizer~\cite{Chandra2022GD} and Early Stopping Against Label Noise Without Validation Data, referred to as Label Wave~\cite{yuan2024early}. A comparison between NeVe and these methods is shown in Tab.~\ref{tab:comparison_related_works}. While these approaches share some similarities with NeVe, they exhibit key differences:
\begin{itemize}
    \item \textbf{Ultimate Optimizer}: Despite its advantages, its performance heavily depends on fine-tuning its hyperparameters. Its effectiveness can vary significantly without precise adjustments or deep stacks of hyperoptimizers. In contrast, NeVe demonstrated that no hyperparameter optimization was necessary when switching between models and tasks.
    \item \textbf{Label Wave}: We evaluated this method using the official implementation provided in the supplementary material of~\cite{yuan2024early}, replacing their ResNet model with the standard ResNet-32 used in our experiments for a fair comparison. Since the paper does not specify the patience value, we set it to 5 and 10 epochs to align with our setup. The main limitation of Label Wave is its early stopping strategy, which halts training prematurely while the model is still learning, significantly reducing performance.
\end{itemize}

\subsection{Ablation Study}
\label{sec:4_ablation}
This section presents ablation studies on key aspects of NeVe, conducted using a ResNet32 model trained on the CIFAR100 dataset, following the same hyperparameter configuration described in Sec.~\ref{sec:experimentres}, unless otherwise specified.

\begin{figure*}[t]
    \centering
    \begin{subfigure}{\columnwidth}
        \centering
        \includegraphics[width=0.88\columnwidth]{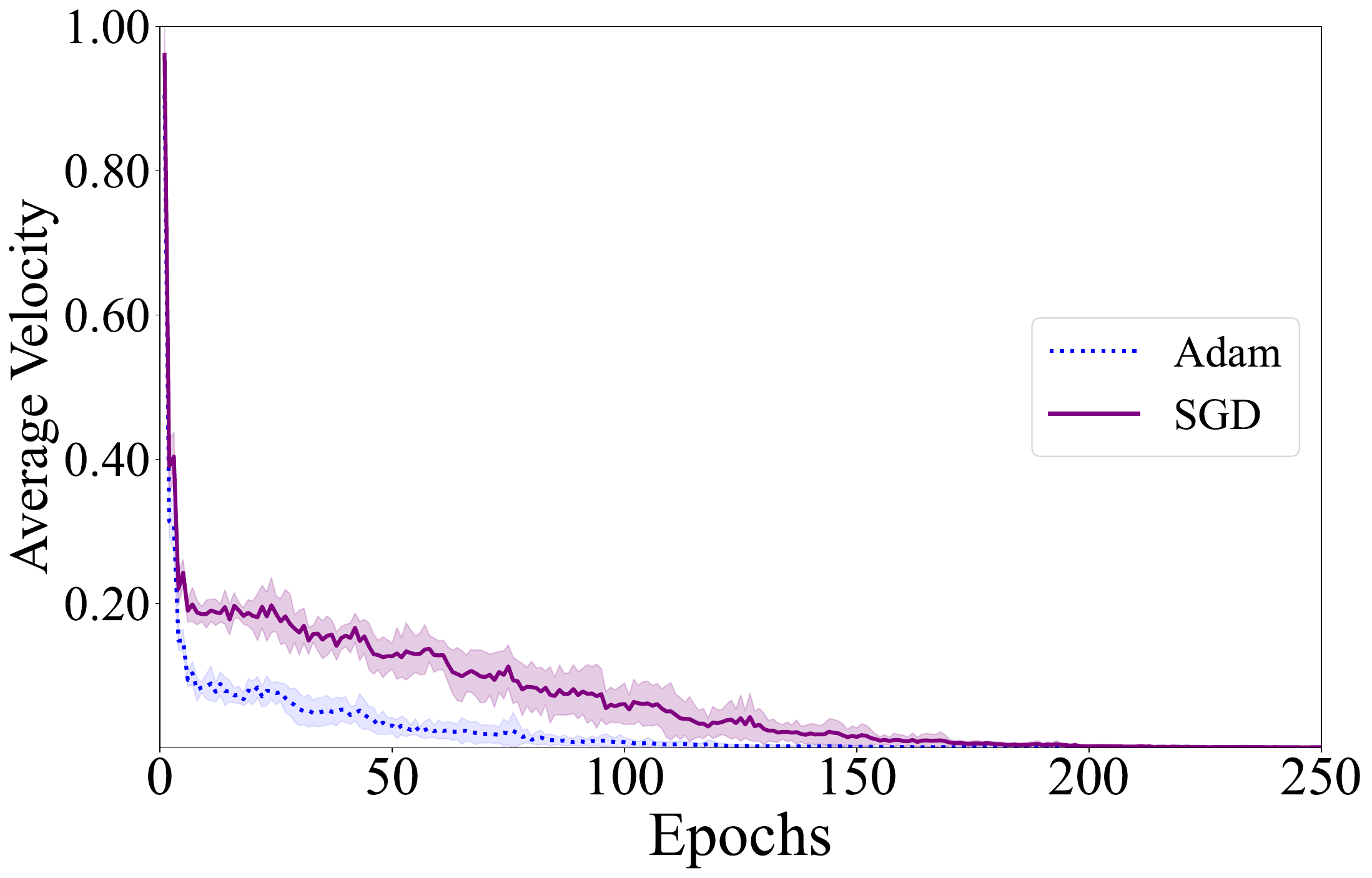}
        \caption{~}
        \label{fig:cap4_adam_vs_sgd_velocity}
    \end{subfigure}
    ~
    \begin{subfigure}{\columnwidth}
        \centering
        \includegraphics[width=0.855\columnwidth]{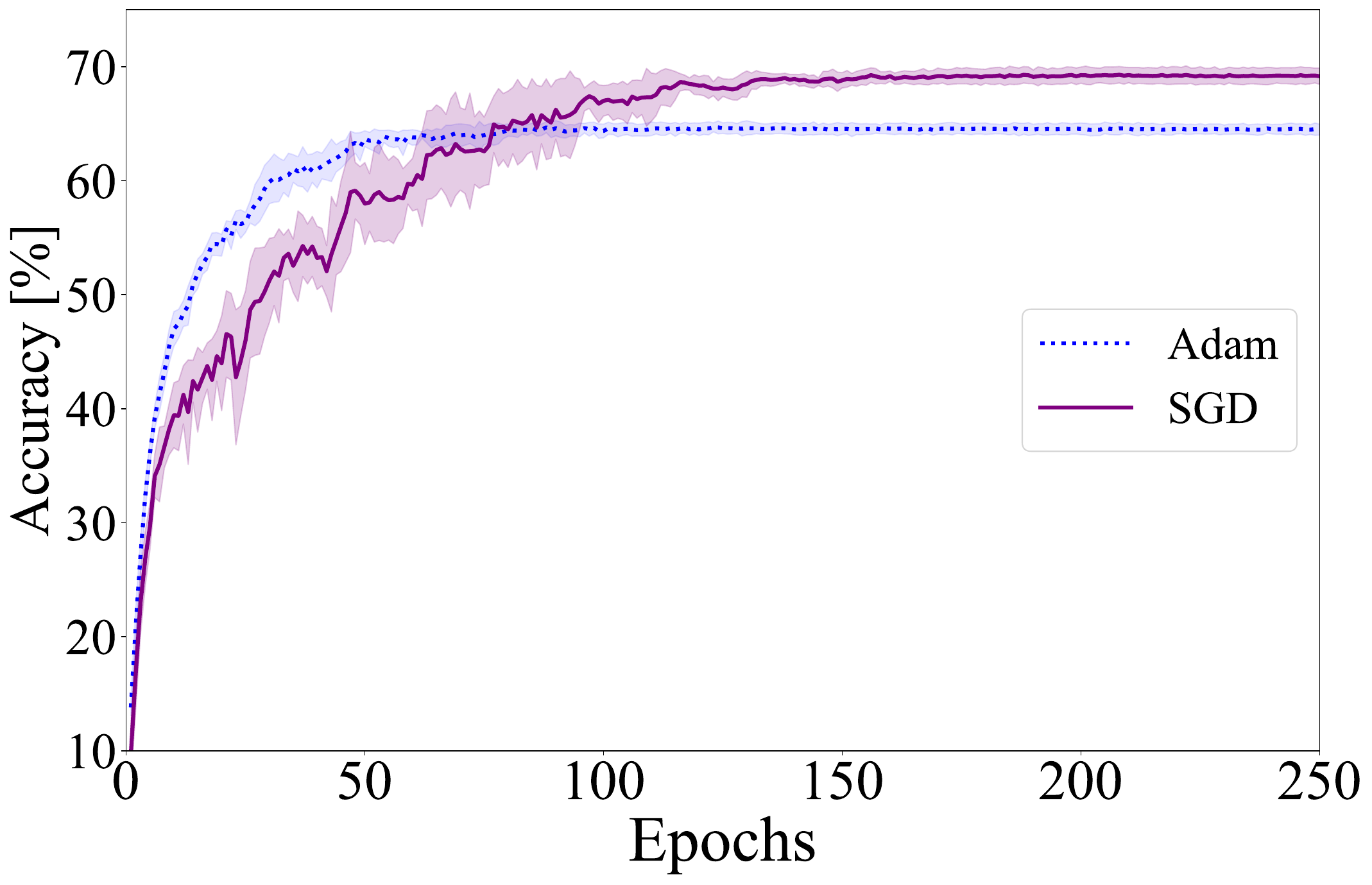}
        \caption{~}
        \label{fig:cap4_adam_vs_sgd_accuracy}
    \end{subfigure}
    \caption{Comparison of NeVe using SGD and Adam in terms of average velocity (a) and test accuracy (b).}
    \label{fig:cap3_ablation_adam_vs_sgd}
\end{figure*}

\textbf{\textit{Adam vs SGD}}  Our first ablation study focuses on the robustness of NeVe to different optimizers. We tested NeVe with SGD and Adam optimizers, changing only the learning rate to $1\times10^{-3}$ for Adam. Fig.~\ref{fig:cap3_ablation_adam_vs_sgd} visualizes the two experiments, averaged on ten seeds. In this simple task, Adam converges to lower accuracy scores ($69.16 \pm 0.71\%$ for SGD and $64.48 \pm 0.48\%$ for Adam after $250$ epochs of training), this is not an unexpected behavior as shown in~\cite{wilson2018marginalvalueadaptivegradient, reddi2019convergenceadam}, still with a faster convergence as showcased by the average velocities (Fig.~\ref{fig:cap4_adam_vs_sgd_velocity}). Furthermore, we observe a similar behavior when applying NeVe’s stopping procedure, with both cases reaching near-to-end performances ($69.18 \pm 0.10\%$ for SGD and $64.54 \pm 0.42\%$ for Adam) in nearly half the training epochs ($148 \pm 20$ for SGD and $140 \pm 24$ for Adam). This showcases the robustness of NeVe to the most popular optimizers.

\input{cap4/main_results}

\textbf{\textit{Stopping Criterion}} Another ablation study focuses on the stopping criteria of NeVe. Namely, we analyze the sensitivity of our velocity-based stopping criterion (where the training ends when $v^t < \varepsilon$) to $\varepsilon$. Fig.~\ref{fig:cap4_ablation_epsilon_accuracy} shows the test accuracy as a function of $\varepsilon$. Here, we observe that a performance plateau occurs when the velocity drops below the threshold $\varepsilon=10^{-3}$, according to what is expected by the formal analysis in Sec.~\ref{sec:3_neve_hyp_estimation}. As a first observation, in Fig.~\ref{fig:cap4_ablation_epsilon_epochs}, we see a decrease in training epochs, as $\varepsilon$ grows. This also corresponds to poorer performance. Among the approaches, for high $\varepsilon$ regimes we observe that $\mathcal{D}_{\text{RND}}$
enables top performance. We believe this is due to the distribution of the tested noise, which allows a more ``task agnostic'' estimation of the velocity.
Ideally, the use of a true validation set should provide a more targeted stopping criterion; however, it is more sensitive to the optimal tuning of $\varepsilon$ (for high values, it behaves the worst), and finding the optimal size of $|\mathcal{D}_{\text{val}}|$ that avoids performance loss is not straightforward.

\subsection{Future Works and Limitations}

In this work, we focused on tuning learning hyperparameters that are typically adjusted during the training process, such as the learning rate and stopping criterion. However, to the present date, it is not known whether NeVe can be exploited to optimize other hyperparameters, such as momentum or weight decay. We hypothesize a direct link between velocity convergence and achieving a plateau in a validation loss curve, supported by our empirical validation. This could allow us to tune the learning rate without needing the parameter $\alpha$; however, more investigations are required. Evaluating the model's convergence opens the doors to further work in the area, such as the design of hyperparameter-free optimization methods in data scarcity scenarios. Moreover, NeVe's stopping criterion could be further improved by introducing adaptive thresholds or incorporating additional metrics to fine-tune the stopping point. A future iteration could integrate task-specific validation objectives based on task-specific metrics. This would enhance NeVe’s ability to make even more informed stopping decisions across a wide range of applications.
To further assess the generality and robustness of the proposed approach, we plan to conduct a comprehensive set of experiments across a diverse array of datasets and tasks. These will include standard benchmarks in image segmentation and object detection, as well as domains beyond computer vision, such as tabular data and other non-visual tasks. We also intend to evaluate NeVe on alternative model architectures, thereby providing additional evidence of its versatility and domain-agnostic nature.
We leave these topics for further research.  Finally, we would like to point out that NeVe is not meant to optimize network design hyperparameters, such as the number of layers or neurons.

%% file: cap4/main_results.tex
\begin{table*}[t]
    \centering
    \caption{Results of the application of NeVe (in bold) compared to the ``Baseline'' and ``V.Loss'' references. We report the performance of the final model as evaluated on the test set classification accuracy. Reported error margins refer to the standard deviation (averaged over 10 runs). $|\mathcal{D}_{\text{val}}|$ is the percentage of samples held out from the training set for validation.}
    \label{tab:cap4_experiments}
    \resizebox{0.66\textwidth}{!}{
    \begin{tabular}{c c c c c c}
        \toprule
        \multirow{1}{*}{\bf Dataset} & \multirow{1}{*}{\bf $\mathbf{|\mathcal{D}_\text{train}|}$} & \multirow{1}{*}{\bf Architecture}&\multirow{1}{*}{\bf Method} & \multirow{1}{*}{\bf $|\mathcal{D}_\text{val}|$} & \multirow{1}{*}{\bf Performance [\%]} \\
        \midrule

        \multirow{4}{*}{CIFAR10} & \multirow{4}{*}{$ 50\text{k} $} & \multirow{4}{*}{ResNet-32} & \multirow{1}{*}{Baseline}
        &  0\% & $ 92.74 \pm 0.34 $ \\
        & & & V.Loss 
           & 30\% & $ 91.38 \pm 0.79 $ \\
        && & V.Loss & 10\% & $ 92.24 \pm 0.48 $\\
        & & & \multirow{1}{*}{\bf NeVe} 
         &\bf 0\% & $ \mathbf{93.10 \pm 0.30} $ \\
        \midrule
        \multirow{4}{*}{CIFAR100} & \multirow{4}{*}{$ 50\text{k} $} & \multirow{4}{*}{ResNet-32} & \multirow{1}{*}{Baseline}
        &  0\% & $ 68.84 \pm 0.46 $ \\
        & & & V.Loss 
           & 30\% & $ 65.96 \pm 0.21 $ \\
        && & V.Loss & 10\% & $ 68.68 \pm 0.15 $\\
        & & & \multirow{1}{*}{\bf NeVe} 
         &\bf 0\% & $ \mathbf{69.18 \pm 0.10} $ \\
        \midrule
        \multirow{9}{*}{ImageNet-100} 
        & \multirow{9}{*}{$ 130\text{k} $} & \multirow{4}{*}{ResNet-50} & \multirow{1}{*}{Baseline}
        &  0\% & $ 82.65 \pm 0.65 $ \\
        && & V.Loss
            & 30\% & $ 82.06 \pm 0.19 $ \\
        && & V.Loss& 10\% & $ 82.70 \pm 0.81 $\\
        && & \multirow{1}{*}{\bf NeVe} 
            &\bf 0\% & $ \mathbf{84.52 \pm 0.51} $ \\
        \cmidrule{3-6}
        & & \multirow{4}{*}{SwinT-v2} &\multirow{1}{*}{Baseline}
        &  0\% & $ 70.05 \pm 0.32 $ \\
        && & V.Loss
            & 30\% & $ 66.67 \pm 0.16 $ \\
        && & V.Loss & 10\% & $ 69.98 \pm 0.23 $ \\
        && & \multirow{1}{*}{\bf NeVe}
            &\bf 0\% & $ \mathbf{69.80 \pm 0.74} $ \\
        \bottomrule
    \end{tabular}
    }
\end{table*}

%% file: cap5/_section_5.tex
\section{Conclusions}
\label{sec:5_conclusion}

This work presented neural velocity as a novel approach to dynamically monitoring and guiding neural network training. By integrating neural velocity into the training process, we developed the NeVe strategy, which adjusts learning rates and stopping conditions without relying on a validation set. Our evaluations show that NeVe performs comparably to traditional training methods across multiple tasks, eliminating the need for extensive tuning and validation data.

The primary advantage of NeVe is its ability to enhance training efficiency, particularly when data is limited, by accurately identifying when further training becomes unnecessary. This reduces computational overhead and resource usage. While NeVe does not directly address model generalization, its method of determining optimal stopping points provides a practical solution for avoiding overfitting, especially in data-constrained scenarios. The implications of this work extend to improving model reliability and training efficiency in various applications where data collection is challenging and expensive.